\documentclass[10pt,twocolumn,letterpaper]{article}

\usepackage{cvpr}              

\usepackage[dvipsnames]{xcolor}

\usepackage{tikz}
\usetikzlibrary{tikzmark}

\usepackage{listings}

\usepackage{graphicx}
\usepackage{amsmath}
\usepackage{amssymb}
\usepackage{xfrac}
\usepackage{color}
\usepackage{colortbl}
\usepackage{enumitem}
\usepackage[dvipsnames]{xcolor}
\newcommand{\Lagr}{\mathcal{L}}
\usepackage{sidecap}
\usepackage{wrapfig}
\usepackage{lipsum} 
\usepackage{textcmds}
\usepackage{graphicx}
\usepackage{multirow}
\usepackage{booktabs}
\usepackage{colortbl}
\usepackage{xcolor}
\usepackage{hhline}
\usepackage{lipsum}
\usepackage{etoolbox}
\usepackage{capt-of,etoolbox}
\definecolor{Gray}{gray}{0.90}
\definecolor{LightCyan}{rgb}{0.82,0.82,1}
\definecolor{tabhighlight}{HTML}{e5e5e5}

\def\etal{\emph{et al.}\xspace}
\def\ie{\emph{i.e.}\xspace}
\def\eg{\emph{e.g.}\xspace}

\usepackage{nicematrix}
\usepackage{listings}
\usepackage{algorithm}
\usepackage{algorithmic}
\usepackage{amsmath,amssymb} %
\usepackage{xcolor,pifont}
\usepackage{standalone}

\newcommand\extrafootertext[1]{%
    \bgroup
    \renewcommand\thefootnote{\fnsymbol{footnote}}%
    \renewcommand\thempfootnote{\fnsymbol{mpfootnote}}%
    \footnotetext[0]{#1}%
    \egroup
}

\definecolor{cvprblue}{rgb}{0.21,0.49,0.74}
\usepackage[pagebackref,breaklinks,colorlinks,citecolor=cvprblue]{hyperref}

\def\paperID{2754} 
\def\confName{CVPR}
\def\confYear{2024}

\title{Generating Action-conditioned Prompts 
for Open-vocabulary  Video \\ Action Recognition}

\author{Chengyou Jia$^{1}$,\;  Minnan Luo$^{1}$,\;  Xiaojun Chang$^{2}$, \; Zhuohang Dang$^{1}$,\; Mingfei Han$^{2}$,\\  Mengmeng Wang$^{3}$, \;Guang Dai$^{4}$,\; Sizhe Dang$^{1}$,\; Jingdong Wang$^{5}$ \\
$^{1}$Xi’an Jiaotong University \; $^{2}$University of Technology Sydney \quad $^{3}$Zhejiang University  \\ $^{4}$SGIT AI Lab \; $^{5}$Baidu Inc\\
\tt\small \{cp3jia, dangzhuohang, darknight1118\}@stu.xjtu.edu.cn \ \tt\small minnluo@xjtu.edu.cn \\
\tt\small \{guang.gdai, cxj273, mhannku030\}@gmail.com  \ \tt\small mengmengwang@zju.edu.cn \ \tt\small wangjingdong@baidu.com
}

\begin{document}
\maketitle
\begin{abstract}




Exploring open-vocabulary video action recognition is a promising venture, which aims to recognize previously unseen actions within any arbitrary set of categories. 
Existing methods typically adapt pretrained image-text models to the video domain, capitalizing on their inherent strengths in generalization. A common thread among such methods is the augmentation of visual embeddings with temporal information to improve the recognition of seen actions. Yet, they compromise with standard less-informative action descriptions, thus faltering when confronted with novel actions. Drawing inspiration from human cognitive processes, we argue that augmenting text embeddings with human prior knowledge is pivotal for open-vocabulary video action recognition. To realize this, we innovatively blend video models with Large Language Models (LLMs) to devise Action-conditioned Prompts. Specifically, we harness the knowledge in LLMs to produce a set of descriptive sentences that contain distinctive features for identifying given actions. Building upon this foundation, we further introduce a multi-modal action knowledge alignment mechanism to align concepts in video and textual knowledge encapsulated within the prompts. Extensive experiments on various video benchmarks, including zero-shot, few-shot, and base-to-novel generalization settings, demonstrate that our method not only sets new SOTA performance but also possesses excellent interpretability.





\end{abstract}    
\section{Introduction}
\label{sec:intro}
\setlength{\belowcaptionskip}{-0.25cm}
\begin{figure}[t]
  \centering
     \centering
     \includegraphics[width=0.47\textwidth]{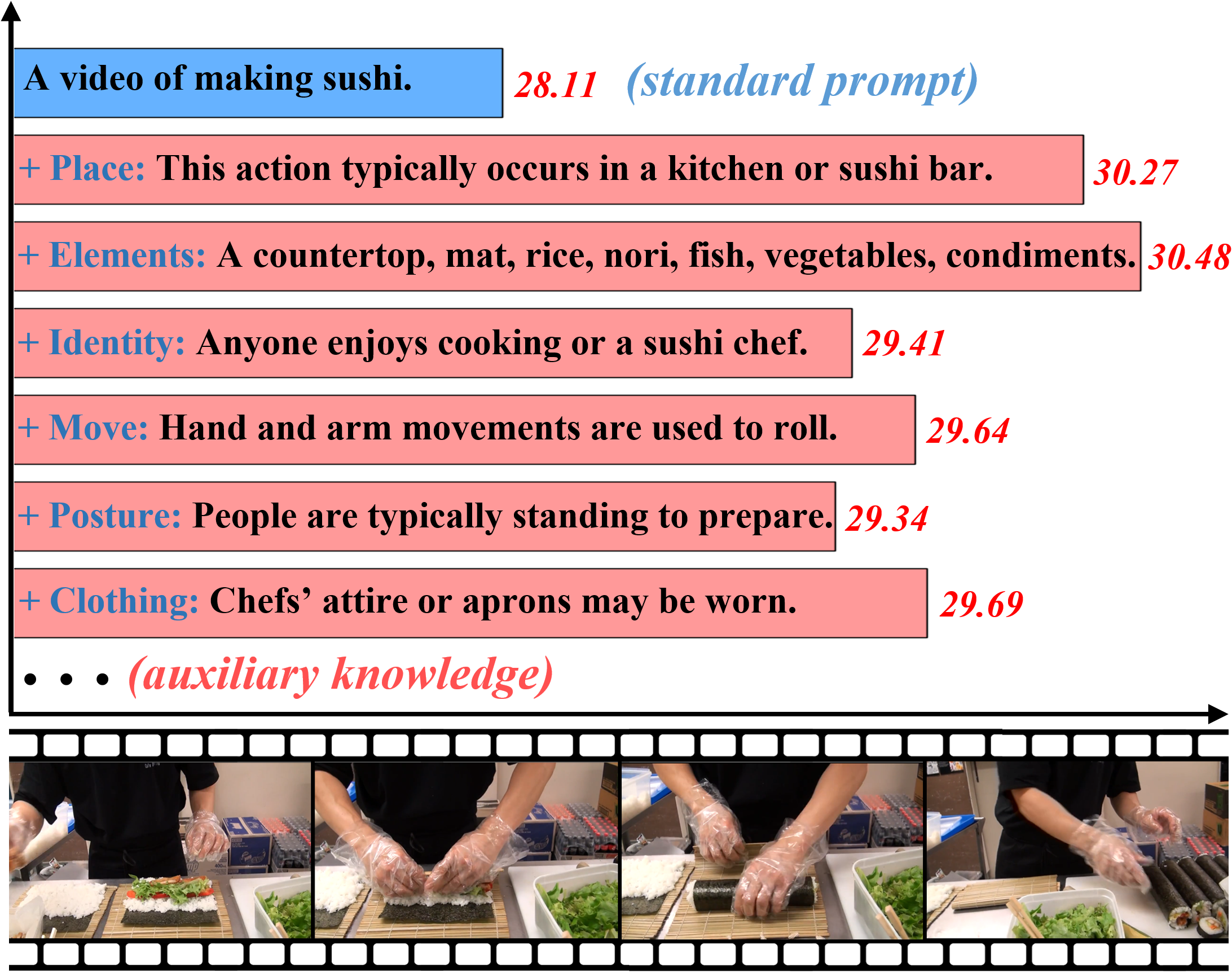}
   \caption{Comparative visualization of prompts and their corresponding match scores. The first line represents a traditional prompt, while the subsequent lines detail prompts that describe the action through multi-attributes. Scores on the right indicate CLIP match scores between the video and textual embeddings of corresponding prompts. It is evident that prompts incorporating auxiliary knowledge significantly enhance the model's recognition.}
   \label{fig:intro}
\end{figure}

\setlength{\belowcaptionskip}{-0.3cm}
\begin{figure}[t]
  \centering
  \includegraphics[width=0.48\textwidth]{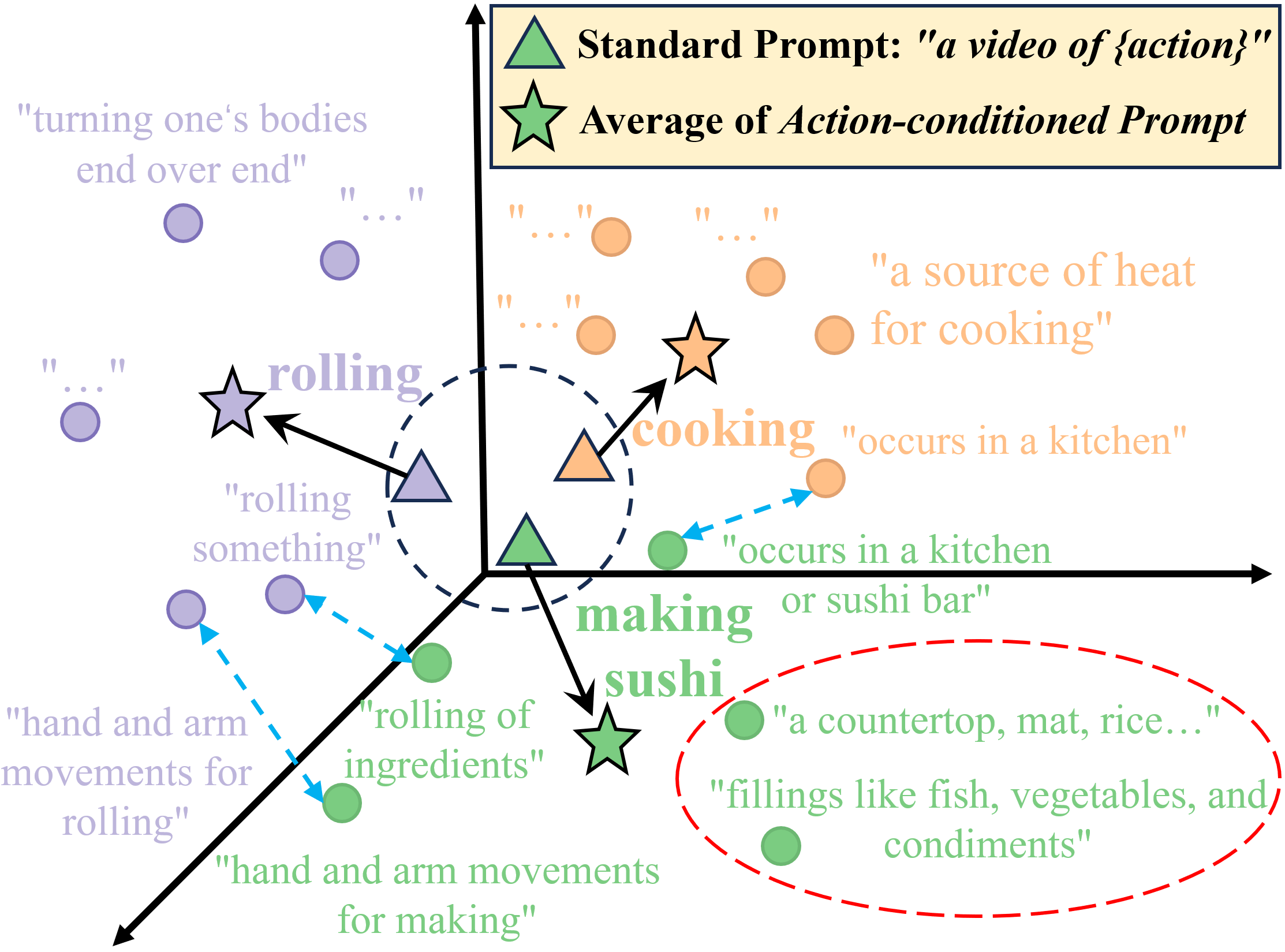}
  \caption{Comparative visualization of text embeddings space for different prompts.  The standard prompt yields limited information, concentrating embeddings within a confined area of the text space, thus proving inadequate for recognizing novel actions. In contrast, Action-conditioned prompts offer multi-attribute descriptions of actions, not only establishing connections (\textit{dashed blue double arrow}) with foundational actions but also providing the necessary knowledge (\textit{dashed red ellipse}) to discern novel actions.}
  \label{fig:adv}
\end{figure}
\setlength{\belowcaptionskip}{-0.0cm}

\textbf{\textit{Why can humans effortlessly recognize novel actions in videos, even with limited or no prior exposure to those specific actions?}} For instance, one can easily identify the action of ``making sushi" in Figure \ref{fig:intro}, despite having rarely witnessed the process of sushi preparation. This remarkable ability primarily arises from two factors. First, a comprehensive understanding of foundational actions allow humans to quickly approximate a novel action, \eg, familiarity with the general action of cooking aids in the recognition of making sushi as a culinary process. Secondly, auxiliary knowledge involved in specific actions serves as supplementary insights to enhance the recognition of novel unseen actions, \eg, recognizing the chef's role, identifying raw fish slices, and capturing the distinctive rolling motion in ``sushi making". 
Drawing inspiration from human cognitive behavior, we posit that an effective open-vocabulary video action model should mirror these two factors, thus empowering it to recognize any arbitrary actions without prior exposure.

Existing approaches \cite{ju2022prompting,wang2021actionclip,ni2022expanding,lin2022frozen} always build open-vocabulary video action models on the foundation of pretrained Vision-Language (VL) models, such as CLIP \cite{clip}. The primary aim is to harness CLIP's robust generalization capabilities and extend them to the video domain \cite{wang2021actionclip}. This is achieved by calculating the similarity between query video and textual embeddings of various categories, with the highest similarity score indicating the matched category.  The open-vocabulary capability stems from the fact that any action categories can be represented and matched through text. Building on this premise, a common thread among existing methods is the integration of temporal modeling to evolve the image encoder into a video encoder, with techniques like cross-frame interactions \cite{ju2022prompting} or temporal attention \cite{ni2022expanding}. These advancements have been instrumental in improving the temporal perception of trained actions, as evidenced in various studies 
\cite{wang2021actionclip,ni2022expanding,pan2022st}.

Nevertheless, such approaches can be regarded as addressing only the first factor. While they excel in recognizing foundational actions, these methods generally fall short when tasked with recognizing novel actions \cite{rasheed2023fine}. In Figure \ref{fig:intro}, we illustrate these methods relying on basic, manually designed prompts such as ``a video of making sushi'' often yield lower CLIP match scores, lacking a clear mechanism to discern novel actions. Considering the second factor mentioned previously, it's crucial to equip the model with additional knowledge pertaining to novel actions. Such considerations should encompass multi-attribute descriptions like the scene of the action, involved elements, relevant props, and so forth. As shown in Figure \ref{fig:intro}, integrating this auxiliary knowledge into the text encoder enables it to anchor the visual content, thereby enhancing the model’s ability to recognize novel actions. Motivated by these observations, our key insight is that distinct actions, especially novel actions that are previously unencountered, should be associated with their own set of knowledge-rich prompts, which we term as \textbf{\textit{Action-conditioned Prompts}}. As shown in Figure \ref{fig:adv}, these prompts are distinct from more generic, hand-written prompts. They not only establish connections with foundational actions but also facilitate the recognition of novel actions through specialized knowledge.

However, manually crafting these prompts presents significant challenges: 1) the process is resource-intensive, both in time and cost, making it impractical for large sets of actions.  2) the variability in annotators' perceptions may result in inconsistent and subjective descriptions, inevitably introducing biases. 
To address these challenges, we blend video action models with Large Language Models (LLMs) to devise Action-conditioned Prompts. Our algorithm begins by systematically identifying 12 pivotal attributes across the categories of Scene, Actor, and Body-related aspects, as depicted in Figure \ref{fig:overview1}. 
We then utilize GPT-4 to generate knowledge-rich descriptions corresponding to each predefined attribute. These descriptions are then synthesized into Action-conditioned Prompts, processed through CLIP's text encoder and combined classifiers to advance open-vocabulary video action recognition. Building on these generating prompts, we further introduce a Multi-modal Action Knowledge Alignment (MAKA) mechanism to align visual concepts in video and textual knowledge within prompts. Specifically, we adopt the cross-modal late interaction, enabling the model to capture the fine-grained relevancy between each prompt and each frame. 


Extensive experiments demonstrate that our method exhibits significant advancements over established baselines in various scenarios including zero-shot, few-shot, and base-to-novel generalization settings, as validated across five distinct video benchmarks. 
In all these extensive settings and metrics, our approach has consistently set new state-of-the-art standards, showcasing its effectiveness and robustness. Moreover, our method possesses excellent interpretability, providing a clear pathway to understanding how the model makes decisions when discerning actions through visual and textual cues.
\section{Related Works}
\label{sec:related}

\subsection{Video Action Recognition}

The realm of video action recognition can be delineated into two principal methodologies: \textit{uni-modal} and \textit{multi-modal} approaches. Uni-modal methods are purely visual models, intensely focused on encoding both spatial and motion cues. Early approaches \cite{christoph2016spatiotemporal,simonyan2014two,wang2016tsn,7780666} leveraged various low-level streams to capture temporal information, \eg, optical flow and RGB differences.
More recently, advanced mechanisms like 3D CNNs ~\cite{carreira2017quo,christoph2016spatiotemporal, feichtenhofer2019slowfast, wang2017spatiotemporal} and video transformers~\cite{arnab2021vivit, liu2022video, yan2022multiview} are proposed to effectively model the long-range spatio-temporal relationships and have shown consistent improvements. 

Complementing these uni-modal methods, the advent of Visual-Language (VL) pre-training \cite{clip} has catalyzed the emergence of multi-modal methods. These innovative multi-modal approaches aim to harness CLIP's generalized VL representations for video recognition. A thread among such methods is the integration of temporal modeling to evolve the image encoder into 
a video encoder. For instance, Ni \etal \cite{ni2022expanding} propose a cross-frame attention mechanism that explicitly exchanges information across frames. Pan \etal \cite{pan2022st} develop 3D convolutional modules as adapters within the CLIP framework. Works such as ActionCLIP \cite{wang2021actionclip}, STAN \cite{liu2023revisiting}, ATM \cite{atm} also adopt similar strategies. Yet, these approaches generally underperform when tasked with identifying novel actions \cite{rasheed2023fine}. In contrast, our work innovatively introduces knowledge-rich action-conditioned prompts, aiming to surpass these limitations and enhance the recognition of novel actions.

\subsection{Prompt Learning using LLM}

The art of prompt engineering holds significant sway in refining the accuracy of language models \cite{brown2020language,gao2020making} and vision-language models \cite{clip}, which has incited extensive research into optimizing prompt formats. Early efforts ranged from assembling manually crafted prompts \cite{bach2022promptsource} to devising learnable prompt tokens \cite{li2021prefix,lester2021power,zhou2021coop,zhou2022conditional}. Advancing beyond these, contemporary studies have leveraged prompts auto-generated by LLMs~\cite{menon2022visual,yang2023language,maniparambil2023enhancing,pratt2023does}, using them to create structured attribute lists that are reformulated into captions for use with CLIP. These methods demonstrate how the rich knowledge embedded in LLMs can effectively augment the perceptual capabilities of visual models. Considering that previous methods have primarily concentrated on fine-grained zero-shot image classification, there remains a lack of a systematic approach for exploring knowledge-rich prompts specifically tailored to actions. This work seeks to complement the scarce literature by introducing innovative Action-conditioned Prompts for video action recognition. 
\setlength{\belowcaptionskip}{-0.35cm}
\begin{figure*}[t]
  \centering
     \centering
     \includegraphics[width=0.9\textwidth]{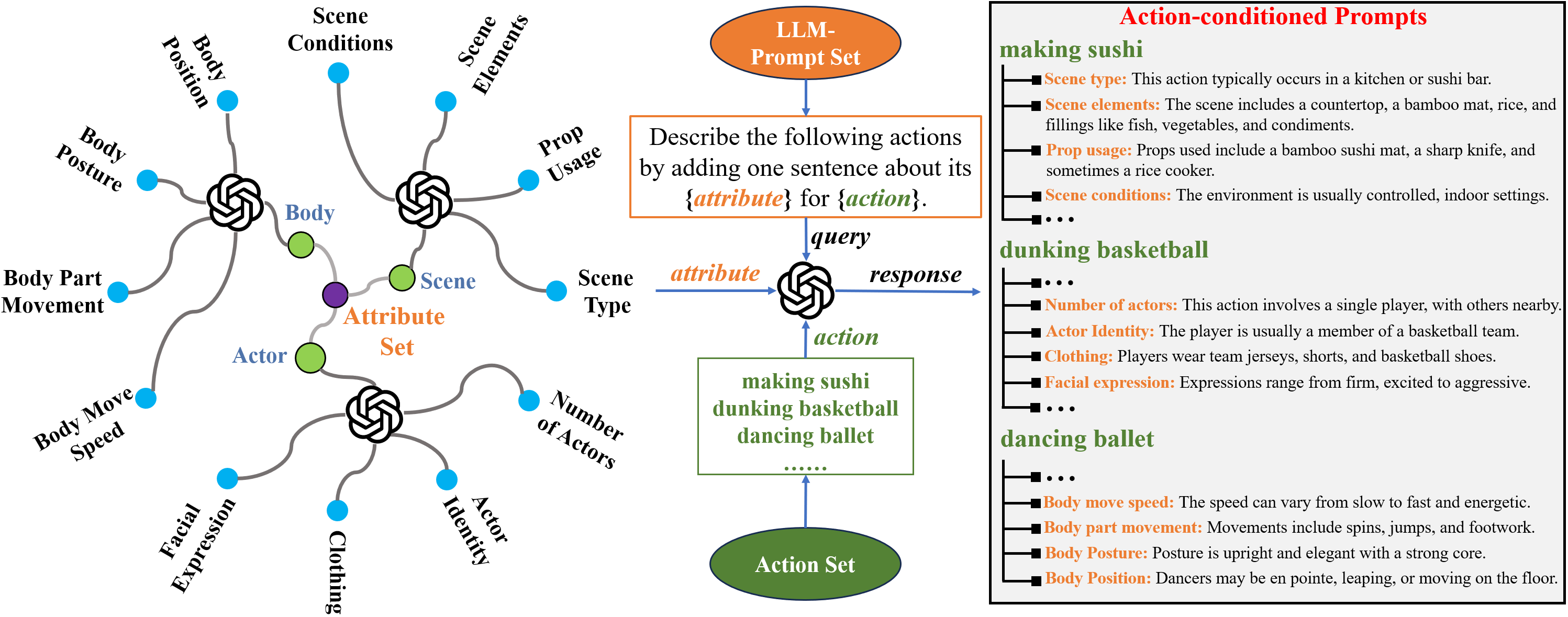}
   \caption{Illustration of Action-conditioned Prompts
 generation workflow: On the left, the process of defining the attribute set is visualized. The middle section depicts the querying process with LLMs, transforming action and attributes into structured prompts. On the right, we present sample snippets of the prompts generated.}
   \label{fig:overview1}
\end{figure*}

\section{Method}
In this section, we first briefly overview the architecture of the CLIP model for video action recognition. Then, we elaborate on the critical component of our approach: Action-conditioned Prompts. Finally, we introduce the Multi-modal Action Knowledge Alignment (MAKA) mechanism, which is designed to enrich video recognition with a more nuanced understanding of actions through generating prompts. 

\subsection{Adapt CLIP for video action recognition}

Given a video $V \in \mathbb{R}^{T \times H \times W \times 3} $ with $T$ frames and a text description $C$, where $V$ and $C$ are sampled from a set of videos and a collection of action category names respectively, we feed the $T$ frames into the video encoder $f_{\theta_v}$ and the text $C$ into the text encoder $f_{\theta_t}$ to obtain a video representation $v$ and a text representation $c$ correspondingly,
\begin{equation}
v = f_{\theta_v}(V), c = f_{\theta_t}(C).  
\end{equation}
The primary objective in fine-tuning clip for video action model lies in maximizing the similarity $sim(v, c)$ if $v$ and $c$ are correspondingly matched, and otherwise minimizing it. Typically, the similarity is calculated using cosine similarity,
\begin{equation}
sim(v, c) = \frac{\langle v, c \rangle}{\|v\| \|c\|}.
\label{equ:sim}
\end{equation}

During inference, the similarity score is calculated between the given video and each action category, with the highest-scoring category being designated as the video's top-1 predicted classification.

\subsection{Action-conditioned Prompts Generation}
Figure \ref{fig:intro} demonstrates the importance of supplementing models with expansive knowledge to augment action recognition. However, acquiring expert annotations is both cost-prohibitive and labor-intensive. It is also subject to individual biases, leading to inconsistent results. To address this, we harness the capabilities of Large Language Models (LLMs), \eg, GPT-4, known for their extensive knowledge and versatility. Despite the absence of visual training data, LLMs can generate descriptions that capture essential action characteristics. This capability stems from the extensive text data used in their training, which is authored by individuals imbued with visual knowledge, indirectly laying a foundation for action recognition.

However, eliciting detailed descriptions for each action directly from LLMs fails to yield satisfactory results. On one hand, hand-written queries usually lead to disparate descriptions, creating a mosaic of focal points across different categories, which in turn induces biases in action recognition. On the other hand, single verbose descriptions hamper the model's interpretability. 
It becomes non-trivial to discern which aspects of the description play a pivotal role in the model's comprehension of actions. 
Therefore, drawing inspiration from LLMs' chain-of-thought \cite{maniparambil2023enhancing,wei2022chain}, we adopt a hierarchical generation strategy to generate multi-attribute Action-conditioned Prompts. 
This strategy entails decomposing the generative task into sequential, manageable stages, tackling each before arriving at the final answer. We then describe the specific steps.

\subsubsection{Which attributes are critical for action recognition?}
To discern critical attributes for action recognition, we initially categorize the components of an action into three fundamental aspects: Scene, Actor, and Body. We then consult GPT-4 to ascertain which attributes within each component are necessary to differentiate actions. From GPT-4's responses, we select the four most representative attributes for each aspect. This process culminates in 12 core attributes, distributed across the three components, as displayed in the left part in Figure \ref{fig:overview1}. For a comprehensive understanding of the inquiries and GPT-4's responses, refer to the appendix. 

\subsubsection{Describe the action about critical attributes}

Building on the attributes identified in the initial stage, we further utilize GPT-4 to generate knowledge-rich descriptive sentences for each action category, forming the basis for our Action-conditioned Prompts. Specifically, we first construct a set of LLM-prompts. 
Then, for each LLM-prompt, we generate a suite of 12 distinct action-conditioned prompts, ensuring that every action is matched with tailored descriptive phrases, as depicted in Figure \ref{fig:overview1}.

Taking the HMDB dataset as an example, which contains 51 categories, if 3 LLM-prompts are used for each category, the total number of prompts generated would be $1886=3\times12\times51$. We constrain each prompt to a maximum of 30 tokens, truncating at the end of a sentence to ensure succinctness. Further refinements include the elimination of superfluous spaces and standardization of punctuation, enhancing consistency. 
Selected examples are showcased in the right segment of Figure \ref{fig:overview1}. Before inputting each prompt into the action recognition model, we concatenate the action-conditioned prompt with a standard prompt format, ``a video of \{action\}'', to explicitly denote the represented action. For more LLM-prompts across all datasets and more detailed design specifics, please refer to the appendix in the supplementary material.

\setlength{\belowcaptionskip}{-0.4cm}
\begin{figure}[t]
  \centering
     \centering
     \includegraphics[width=0.48\textwidth]{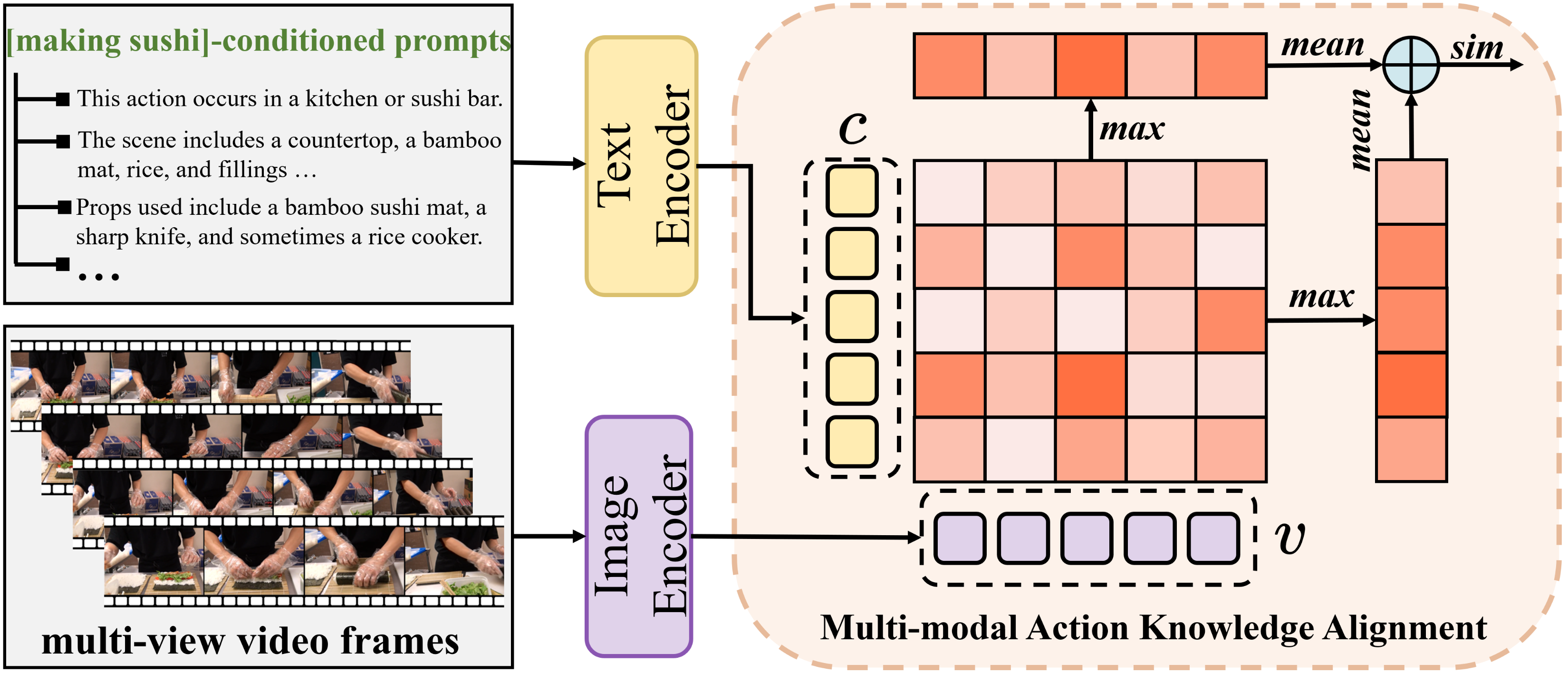}
   \caption{Illustration of the multi-modal action knowledge alignment mechanism.}
   \label{fig:overview2}
\end{figure}
\setlength{\belowcaptionskip}{-0.0cm}

\subsection{Multi-modal Action Knowledge Alignment}
These generated action-conditioned prompts equip the model with a multifaceted understanding of actions. However, the challenge that follows is effectively aligning these prompts, which offer various perspectives, with corresponding visual concepts within the videos. Previous methods temporally aggregate the embeddings of video frames, \eg, mean pooling in \cite{rasheed2023fine} or attention pooling in \cite{ni2022expanding}, to yield a video-level representation. 
This approach, while prevalent, is not conducive to fine-grained concept alignment and leaves much to be desired in terms of nuanced video representation and alignment strategies. To tackle this issue, we introduce a novel \textbf{Multi-modal Action Knowledge Alignment} mechanism to foster a more precise correspondence between text and video data.

To capture the multi-faceted features of videos, we implement a multi-view strategy \cite{carreira2017quo,feichtenhofer2019slowfast,wu2021mvfnet} that samples multiple clips per video with several spatial crops, which allows the pre-trained vision-language model to encode each video into multiple frame embeddings, as illustrated in Figure \ref{fig:overview2}. Further, inspired by \cite{khattab2020colbert,yao2021filip}, we apply a cross-modal late interaction to model the fine-grained relevancy between each prompt and each frame. 

Specifically, 
we define $n_v$ and $n_t$ as the count of frames for the video $V$ and the number of prompts for the category $C$, respectively. 
The visual features, denoted as $v = f_{\theta_v}(V) \in \mathbb{R}^{n_v \times d}$, and the prompt embeddings, denoted as $c = f_{\theta_t}(C) \in \mathbb{R}^{n_t \times d}$, are encoded accordingly. Here, 
$v$ and $c$ represent the normalized embeddings. Note that the current $C$ no longer refers to the category name but rather a series of action-conditioned prompts.  The fine-grained similarity between $v$ and $c$ is computed via the following process:

For the $i$-th visual features in $v$, we assess its similarity across all prompt embeddings $c$, selecting the highest similarity by
\begin{equation}
\max_{0\le j < n_t} v_i^\top c_j,
\label{eq:tokenwise_max_sim}
\end{equation} 
which represents the maximum fine-grained similarity within the $C$ category.
The video-to-category similarity is then the average of these maximum similarities across all visual features, given by
\begin{equation}
sim_{v2t}(v,c)  = \frac{1}{n_v}\sum_{i=1}^{n_v} \max_{0\le j < n_t} v_i^\top c_j. \label{eq:late_sim_i}
\end{equation}
Conversely, for each $i$-th prompt embedding in $c$, we calculate its similarity with all visual features $v$, adopting the highest as the fine-grained maximum similarity. The category-to-video similarity is the average of these values,
\begin{equation}
sim_{t2v}(v,c)  = \frac{1}{n_t}\sum_{i=1}^{n_t} \max_{0\leq j < n_v} v_j^\top c_i. \label{eq:late_sim_t}
\end{equation}

We refine the cosine similarity computation (Equation \eqref{equ:sim}), in the training and inference processes by integrating both video-to-category and category-to-video similarities, 
\begin{equation}
sim(v,c)  = \frac{1}{2} (sim_{v2t}(v,c) + sim_{t2v}(v,c)).
\label{equ:sim_f}
\end{equation}
For model fine-tuning, we utilize a standardized protocol as outlined in ViFi-CLIP \cite{rasheed2023fine}. More details of the training objectives and procedures are provided in the appendix. 

Though inspired from \cite{khattab2020colbert,yao2021filip}, it diverges significantly in its application of the cross-modal late interaction. Prior methods \cite{yao2021filip} have concentrated on fine-grained image-text matching at a token level, necessitating extensive training to relearn these associations. Such an approach does not contribute to enhancing generalizability. In contrast, the primary aim of our method is to establish frame-to-prompt correspondences, grounding similarity pairings in the natural image-text matching capabilities that are a forte of CLIP.  By leveraging this inherent strength, frame-to-prompt associations are more conducive to improved generalization.


\section{Experiments}
\label{sec:exp}

\subsection{Experimental Setup}

\noindent \textbf{Datasets.} We conduct experiments on five widely used video benchmarks: Kinetics-400 and 600~\cite{k400, k600}, HMDB-51~\cite{kuehne2011hmdb}, UCF-101~\cite{soomro2012ucf101} and SSv2~\cite{goyal2017ssv2}. See the appendix for more details of these datasets.

\noindent \textbf{Implementation details.}  
We use ViT-B/16 based CLIP \cite{clip} model for our experiments. Our adaptation of the CLIP model follows \cite{rasheed2023fine}, with tailored modifications to the prompts and fine-grained similarity function used. We refer to our method as AP-CLIP (\textit{\textbf{A}ction-conditioned \textbf{P}rompt}). We use GPT-4 as the default prompt generator, and the default number of LLM-prompts is 3. Moreover, we align with previous methods \cite{wang2021actionclip,ni2022expanding,ju2022prompting,rasheed2023fine} for various settings including zero-shot, base-to-novel, few-shot, and fully-supervised. Specifically, we utilize 8 frames and employ multi-view inference incorporating 2 spatial crops and 2 temporal views. In the fully supervised setting, our approach extends to using 16 frames, combined with multi-view inference featuring 4 spatial crops and 3 temporal views, consistent with compared methods. For a comprehensive understanding, more detailed prompts and training configurations are provided in the appendix.

\begin{table}[!t]
\centering
\caption{\textbf{Zero-shot setting:} We compare our AP-CLIP with uni-modal and CLIP-based approaches. Besides our AP-CLIP, generating action-conditioned prompts can be seamlessly integrated into the zero-shot approach, Open-VCLIP \cite{weng2023transforming}.  Gains over previous methods are indicated in \textcolor{MidnightBlue}{blue}. Methods marked with `*' are re-evaluated using their official code. }

\setlength{\tabcolsep}{6.5pt}
\scalebox{0.85}{
\begin{tabular}{lccc} \toprule
    Method  &  HMDB-51 & UCF-101 & K600 \\ \midrule
    \midrule
    \rowcolor{tabhighlight}\multicolumn{4}{c}{Uni-modal zero-shot action recognition models} \\
    \midrule
    ASR ~\cite{wang2017alternative} &  21.8 $\pm$ 0.9 & 24.4 $\pm$ 1.0 & - \\
    ZSECOC ~\cite{qin2017zero}  &   22.6 $\pm$ 1.2 & 15.1 $\pm$ 1.7  & -\\
    UR ~\cite{zhu2018towards} & 24.4 $\pm$ 1.6 & 17.5 $\pm$ 1.6 & -\\
    E2E ~\cite{brattoli2020rethinking}  & 32.7 & 48 & -\\
    GCN~\cite{ghosh2020all}  & - & - & 22.3 $\pm$ 0.6  \\
    ER-ZSAR ~\cite{chen2021elaborative}  & 35.3 $\pm$ 4.6 & 51.8 $\pm$ 2.9 & 42.1 $\pm$ 1.4  \\
    \midrule
    \rowcolor{tabhighlight}\multicolumn{4}{c}{Adapting pre-trained VL models (\textit{ViT-B/16})} \\
    \midrule
    Vanilla CLIP~\cite{clip} & 40.8 $\pm$ 0.3 & 63.2 $\pm$ 0.2  & 59.8 $\pm$ 0.3\\
    ActionCLIP~\cite{wang2021actionclip} & 40.8 $\pm$ 5.4 & 58.3 $\pm$ 3.4 &  66.7 $\pm$ 1.1\\
    XCLIP~\cite{ni2022expanding} &  44.6 $\pm$ 5.2 & 72.0 $\pm$ 2.3 & 65.2 $\pm$ 0.4\\
    A5~\cite{ju2022prompting} & 44.3 $\pm$ 2.2 & 69.3 $\pm$ 4.2 & 55.8 $\pm$0.7 \\
    
    VicTR \cite{kahatapitiya2023victr} & 51.0 $\pm$ 1.3 & 72.4 $\pm$ 0.3 & - \\
    ViFi-CLIP \cite{rasheed2023fine}       & 51.3 $\pm$ 0.6 & 76.8 $\pm$ 0.7 & 71.2 $\pm$ 1.0  \\
    AP-CLIP(ours)  & \textbf{55.4 $\pm$ 0.8} & \textbf{82.4 $\pm$ 0.5} & \textbf{73.4 $\pm$ 1.0}  \\
     & {\textcolor{MidnightBlue}{{+4.1}}} &  {\textcolor{MidnightBlue}{{+5.6}}} &  {\textcolor{MidnightBlue}{{+2.2}}}\\
    \midrule
    Open-VCLIP  \cite{weng2023transforming} & 53.9 $\pm$ 1.2 & 83.4 $\pm$ 1.2 & 73.0 $\pm$ 0.8  \\
    \textbf{\textit{+ Action prompts}} & \textbf{57.0 $\pm$ 0.8} & \textbf{85.1 $\pm$ 1.2} & \textbf{74.4 $\pm$ 0.7}  \\
    & {\textcolor{MidnightBlue}{{+3.1}}} &  {\textcolor{MidnightBlue}{{+1.7}}} &  {\textcolor{MidnightBlue}{{+1.4}}}\\
    \midrule
    \rowcolor{tabhighlight}\multicolumn{4}{c}{Adapting pre-trained VL models (\textit{ViT-L/14})} \\
    BIKE* \cite{wu2023bidirectional}  & 50.2 $\pm$ 3.7 & 79.1 $\pm$ 3.5 & 68.5 $\pm$ 1.2  \\
    Text4Vis \cite{wu2022transferring}  & 58.1 $\pm$ 5.7 & 85.8 $\pm$ 3.3 & 68.9 $\pm$ 1.0  \\
    Open-VCLIP  \cite{weng2023transforming} & 59.0 $\pm$ 0.6 & 87.6 $\pm$ 1.2 & 81.1 $\pm$ 0.8  \\
    
    \textbf{\textit{+ Action prompts}} & \textbf{60.0 $\pm$ 1.4} & \textbf{90.2 $\pm$ 0.4} & \textbf{81.9 $\pm$ 1.0} \\
    & {\textcolor{MidnightBlue}{{+1.0}}} &  {\textcolor{MidnightBlue}{{+2.6}}} &  {\textcolor{MidnightBlue}{{+0.8}}}\\

\bottomrule
\end{tabular}
}
\vspace{-0.3cm}
\label{p1_zeroshot}

\end{table}

\begin{table*}[t]
\centering
\caption{\textbf{Base-to-novel generalization:} We compare the generalization ability of AP-CLIP with other models that adapt CLIP~\cite{clip}. The values to the left of the ``/'' symbol indicate that models commence training from the native parameters of CLIP, while the right values denote that models are initially pre-trained on Kinetics-400, serving to bridge the modality gap. HM refers to harmonic mean which measures the trade-off between base and novel accuracy. Gains are shown in \textcolor{MidnightBlue}{blue}.}
\vspace{-0.15cm}
\setlength{\tabcolsep}{3.25pt}
\scalebox{0.81}{
\begin{tabular}[t]{l ccc|ccc|ccc|ccc}
\toprule
&      \multicolumn{3}{c}{K-400}    &\multicolumn{3}{c}{HMDB-51}  &\multicolumn{3}{c}{UCF-101}    & \multicolumn{3}{c}{SSv2} \\
\cmidrule(lr){2-13}
Method  & Base & Novel & HM          & Base & Novel & HM         & Base & Novel & HM         & Base & Novel & HM   \\
\midrule
Vanilla CLIP~\cite{clip}      & 62.3 & 53.4 & 57.5        & 53.3 / - &  46.8 / - & 49.8 / -       & 78.5 / - & 63.6 / - & 70.3 / -       & 4.9 / - & 5.3 / - & 5.1 / - \\ 
ActionCLIP~\cite{wang2021actionclip}        & 61.0 & 46.2 & 52.6         & 69.1 / 69.0 & 37.3 / 57.2 & 48.5 / 62.6        & 90.1 / 85.6 & 58.1 / 75.3 & 70.7 / 80.1       &  13.3 / 8.1 &  10.1 / 8.7 &  11.5 / 8.4 \\
XCLIP~\cite{ni2022expanding}              &  74.1 &  56.4 &  64.0        &  69.4 / 75.8 & 45.5 / 52.0 &  55.0 / 61.7     & 89.9 / 95.4 &  58.9 / 74.0 &  71.2 / 83.4       & 8.5 / 14.2 & 6.6 / 11.0 & 7.4 / 12.4 \\
A5~\cite{ju2022prompting}     & 69.7 & 37.6 & 48.8        & 46.2 / 70.4 & 16.0 / 51.7 & 23.8 / 59.6       &  90.5 / 95.8 & 40.4 / 71.0 & 55.8 / 81.6       & 8.3 / 12.9 & 5.3 / 5.7 & 6.4 / 7.9 \\
 
ViFi-CLIP \cite{rasheed2023fine}     & 76.4 & 61.1 & 67.9         & 73.8 / \textbf{77.1} & 53.3 / 54.9 & 61.9 / 64.1        & 92.9 / \textbf{95.9} & 67.7 / 74.1 & 78.3 / 83.6       & 16.2 / 15.8 & 12.1 / 11.5 & 13.9 / 13.3 \\
\midrule
AP-CLIP(ours)    & \textbf{77.2} & \textbf{64.1} & \textbf{70.0}         & \textbf{74.6} / 75.4 & \textbf{55.9} / \textbf{60.3} & \textbf{63.9} / \textbf{67.0}        & \textbf{94.8} / 95.0 &\textbf{77.0 / 82.9}  & \textbf{84.8} / \textbf{88.5}       & \textbf{16.3 / 16.5} & \textbf{12.9 /12.7} & \textbf{14.4 / 14.3}\\

                  &  \textcolor{MidnightBlue}{{+0.8}} &  \textcolor{MidnightBlue}{{+3.0}}  &  \textcolor{MidnightBlue}{{+2.1}}  &  \textcolor{MidnightBlue}{{+0.8 / -1.7}}  &  \textcolor{MidnightBlue}{{+2.6 / +3.1}}  &  \textcolor{MidnightBlue}{{+2.0 / +2.9}}  &  \textcolor{MidnightBlue}{{+1.7 / -0.9}}  &  \textcolor{MidnightBlue}{{+9.3 / +7.6}}  &  \textcolor{MidnightBlue}{{+6.5 / +4.9}}  &  \textcolor{MidnightBlue}{{+0.1 / +0.7}}  &  \textcolor{MidnightBlue}{{+0.7 / +1.2}}  &  \textcolor{MidnightBlue}{{+0.5 / +1.0}} \\ 

\midrule

\end{tabular}
}
\vspace{-0.2cm}
\label{base_novel}
\end{table*}

\begin{table*}[t]
\centering
\caption{\textbf{Few-shot setting:} The values on two sides of the ``/'' have the same meaning as in Table 2, one from CLIP's native parameters and the other pre-trained on Kinetics-400. Gains are indicated in \textcolor{MidnightBlue}{blue}.}
\vspace{-0.15cm}
\setlength{\tabcolsep}{7.5pt}
\scalebox{0.81}{
\begin{tabular}{l cccc|cccc}
  \toprule
  \multirow{2}{*}{Model} & \multicolumn{4}{c}{HMDB-51} & \multicolumn{4}{c}{UCF-101} \\
  \cmidrule(lr){2-9}
  & $K$=2 & $K$=4 & $K$=8 & $K$=16 & $K$=2 & $K$=4 & $K$=8 & $K$=16 \\
  \midrule
  Vanilla CLIP~\cite{clip} & 41.9 / - & 41.9 / - & 41.9 / - & 41.9 / - & 63.6 / - & 63.6 / - & 63.6 / - & 63.6 / - \\
  ActionCLIP~\cite{wang2021actionclip} & 47.5 / 54.3 & 57.9 / 56.2 & 57.3 / 59.3 & 59.1 / 66.1 & 70.6 / 76.7 & 71.5 / 80.4 & 73.0 / 87.6 & 91.4 / 91.8 \\
  XCLIP~\cite{ni2022expanding} & 53.0 / 60.5 & 57.3 / \textbf{66.8} & 62.8 / 69.3 & 64.0 / 71.7 & 48.5 / 89.0 & 75.6 / 91.4 & 83.7 / 94.7 & 91.4 / 96.3 \\
  A5~\cite{ju2022prompting} & 39.7 / 46.7 & 50.7 / 50.4 & 56.0 / 61.3 & 62.4 / 65.8 & 71.4 / 76.3 & 79.9 / 84.4 & 85.7 / 90.7 & 89.9 / 93.0 \\
  ViFi-CLIP~\cite{rasheed2023fine} & 57.2 / 63.0 & 62.7 / 65.1 & 64.5 / 69.6 & 66.8 / 72.0 & 80.7 / 91.0 & 85.1 / 93.7 & 90.0 / 95.0 & 92.7 / 96.4 \\
  \midrule
  AP-CLIP(ours) & \textbf{59.9} / \textbf{65.1} & \textbf{64.8} / \textbf{66.8} & \textbf{66.8} / \textbf{70.9} & \textbf{68.5} / \textbf{72.5} & \textbf{84.9} / \textbf{92.9} & \textbf{89.1} / \textbf{95.0} & \textbf{91.7} / \textbf{95.8} & \textbf{94.1} / \textbf{96.9} \\
  & \textcolor{MidnightBlue}{{+2.7 / +2.1}} & \textcolor{MidnightBlue}{{+2.1 / +0.0}} & \textcolor{MidnightBlue}{{+2.3 / +1.3}} & \textcolor{MidnightBlue}{{+1.7 / +0.5}} & \textcolor{MidnightBlue}{{+4.2 / +1.9}} & \textcolor{MidnightBlue}{{+4.0 / +1.3}} & \textcolor{MidnightBlue}{{+1.7 / +0.8}} & \textcolor{MidnightBlue}{{+1.4 / +0.5}} \\
  \bottomrule                             
\end{tabular}
}
\vspace{-0.3cm}
\label{p1_few_shot}
\end{table*}

\subsection{Comparisons with State-of-the-art}

\subsubsection{AP-CLIP Generalizes Well !}

To demonstrate the open-vocabulary recognizing capabilities, we follow the benchmark in \cite{rasheed2023fine}, evaluating models in two distinct settings: 1) \textbf{zero-shot setting} and 2) \textbf{base-to-novel setting}. The former primarily assesses the model's capacity to recognize novel actions across different datasets, while the latter tests its performance to recognize novel and rarer actions within the given dataset. Further details regarding these settings can be found in the appendix. 

\noindent\textbf{{(\romannumeral 1) Zero-shot Setting}}: We train the model on a large video action recognition dataset, Kinetics-400 and evaluate across different datasets, HMDB-51, UCF-101 and Kinetics-600. Results are presented in Table~\ref{p1_zeroshot}, where our model, AP-CLIP, is benchmarked against both uni-modal methods and other CLIP-based approaches. It's evident that even the vanilla CLIP demonstrates an impressive generalization performance as compared to uni-modal methods. Further analysis reveals that methods like ActionCLIP and XCLIP, which integrate additional temporal modules, may overfit on trained actions, thereby failing to show substantial generalization improvements. An alternative strategy, exemplified by ViFi-CLIP, involves merely fine-tuning the foundational CLIP model without incorporating external modules, yielding more promising generalization performance.  Against this backdrop, our AP-CLIP also employs a straightforward fine-tuning of the CLIP model, incorporating \textit{action-conditioned prompts} and the \textit{multi-modal action knowledge alignment}. Table~\ref{p1_zeroshot} demonstrates that AP-CLIP yields consistent performance improvements, with gains of +4.1\%, +5.6\%, and +2.2\% on the HMDB-51, UCF-101, and Kinetics-600 datasets, respectively.

Furthermore, our method has demonstrated its adaptability. As a representative, we choose the current best competitor, Open-VCLIP \cite{weng2023transforming}, a robust model specifically designed for zero-shot action recognition. By integrating our action-conditioned prompts in place of its manual prompts, Open-VCLIP experienced a remarkable boost in its generalization capabilities, all without necessitating any retraining. Remarkably, under the ViT-L/14 CLIP model, this integration enhances Open-VCLIP to achieve groundbreaking performance, recording impressive scores of \textbf{60.0\%}, \textbf{90.2\%}, and \textbf{81.9\%} across the three datasets, thereby establishing new state-of-the-art in zero-shot action recognition. These results underscore significant generalization improvements from our action-conditioned prompts.


\noindent\textbf{{(\romannumeral 2) Base-to-novel Generalization Setting}}: In Table~\ref{base_novel}, we evaluate the generalization from base to novel classes on four datasets, K-400, HMDB-51, UCF-101 and SSv2. All methods were initially trained on well-established base classes, while the novel classes represented a realm of previously unencountered scenarios, \ie,  base and novel classes are disjoint. We adopted two distinct approaches for the latter three datasets: one leveraging the original CLIP parameters, and another utilizing parameters pre-trained on Kinetics-400. As shown in Table~\ref{base_novel}, AP-CLIP demonstrates noticeable gains in novel accuracy. Despite observing marginal reductions in base accuracy under certain conditions, our approach effectively balanced the trade-off between base and novel class performance, securing the highest overall harmonic mean on all datasets. We also observed varied gains across different datasets. Temporally challenging datasets like SSv2~\cite{goyal2017ssv2} showed limited improvements, whereas less temporally complex datasets like UCF~\cite{soomro2012ucf101} exhibited significant gains. We look forward to future work focusing on this phenomenon.

\subsubsection{AP-CLIP Specializes Well !}
Our investigation extends to AP-CLIP’s efficacy in narrowing the domain gap within supervised video action recognition tasks. We evaluate its performance under two distinct data availability scenarios:  1) \textbf{few-shot setting}, where the number of training samples is limited, and 2) \textbf{fully-supervised setting}, where we have an abundance of training samples. These settings help us to comprehensively understand and evaluate the specialization performance of our approach under varying levels of data availability.

\noindent\textbf{{(\romannumeral 1) Few-shot Setting}}: Table~\ref{p1_few_shot} delineates AP-CLIP's performance within a few-shot learning scenario, in comparison with other CLIP-based methodologies. AP-CLIP consistently exhibits performance improvements with increasing shots. Across both HMDB-51 and UCF-101 datasets, AP-CLIP surpasses all competing methods in each shot division (2, 4, 8, 16 shots). Notably, the advantage of our approach is more pronounced when training data is scant. The lesser training data provided, the more significant the improvement brought by our method. This suggests that our action-conditioned prompts provide more extensive knowledge about actions, enabling the model to rapidly gain a deeper understanding of actions even with fewer examples.

\noindent\textbf{{(\romannumeral 2) Fully-supervised Setting}}: We compare the performance of AP-CLIP trained on Kinetics-400 with uni-modal video-specific models and other CLIP-based methods in Table~\ref{p1_fullysupervised}. To ensure a fair comparison, results from methods employing ViT-L/14 have been excluded, with all CLIP-related models in this study based on the ViT-B/16 architecture. Despite AP-CLIP's primary intent for novel action recognition, Table~\ref{p1_fullysupervised} indicates its commendable applicability to fully-supervised tasks. Although it may not outstrip the more temporally intricate methods like UniFormerV2~\cite{li2022uniformerv2} or STAN~\cite{liu2023revisiting}, the margin of difference is not significant. Relative to baseline CLIP model fine-tuning \cite{rasheed2023fine}, our approach delivers competitive performance in recognizing trained actions. This substantiates our approach's utility in narrowing the domain gap between image and video modalities.

\begin{table}[t]
	\centering\setlength{\tabcolsep}{4pt}
 \caption{\textbf{Fully-supervised setting:} We compare ViFi-CLIP with uni-modal methods and models specifically designed to adapt CLIP for video tasks on Kinetics-400.}
	\scalebox{0.95}{
		\begin{tabular}{lcccc}
			\toprule
			Method & Frames & Top-1 &  Top-5 & Views  \\ \midrule
			\rowcolor{tabhighlight} \multicolumn{5}{c}{{Uni-modal action recognition models}} \\
			\midrule
			Uniformer-B~\cite{li2022uniformer}   &   32   &   83.0 & 95.4 &  4 $\times$ 3   \\
			TimeSformer-L~\cite{timesformer2021} &   96   &   80.7 & 94.7 &  1 $\times$ 3   \\
			Mformer-HR~\cite{patrick2021keeping} &   16   &   81.1 & 95.2 &  10 $\times$  3   \\
			Swin-L~\cite{liu2021video}    		 &   32   &   83.1 & 95.9 &  4 $\times$ 3 \\

             ViViT-H \cite{arnab2021vivit} &   16   &   84.8 & 95.8 &  4 $\times$ 3   \\

             UniFormerV2-B \cite{li2022uniformerv2} &   8   &   \textbf{85.6} & \textbf{97.0} &  4 $\times$ 3   \\
			\midrule
			\rowcolor{tabhighlight} \multicolumn{5}{c}{{Adapting pre-trained image VL models}} \\
			\midrule
		    ActionCLIP~\cite{wang2021actionclip} & 32 &   83.8  & 96.2 &  10 $\times$ 3   \\
		    X-CLIP~\cite{ni2022expanding} & 16 & 84.7 &  96.8 & 4 $\times$ 3   \\
			A6~\cite{ju2022prompting}  & 16 &  76.9  & 93.5  &  -  \\
                STAN~\cite{liu2023revisiting} & 16 & \textbf{84.9} &  \textbf{96.8}  &  1 $\times$ 3   \\
            \midrule
			ViFi-CLIP \cite{rasheed2023fine}   	&  16  &   83.9  &  96.3 &  4 $\times$ 3  \\
   
                AP-CLIP(ours)  	&  16  &   \textbf{84.7}  &  \textbf{96.7} &  4 $\times$ 3  \\

                &&\textcolor{MidnightBlue}{{+0.8}}  &  \textcolor{MidnightBlue}{{+0.4}} & \\
			\bottomrule

		\end{tabular}}
	\label{p1_fullysupervised}
	\vspace{-0.25in}
\end{table}

\setlength{\belowcaptionskip}{-0.4cm}
\begin{figure*}[t]
  \centering
  \includegraphics[width=0.99\textwidth]{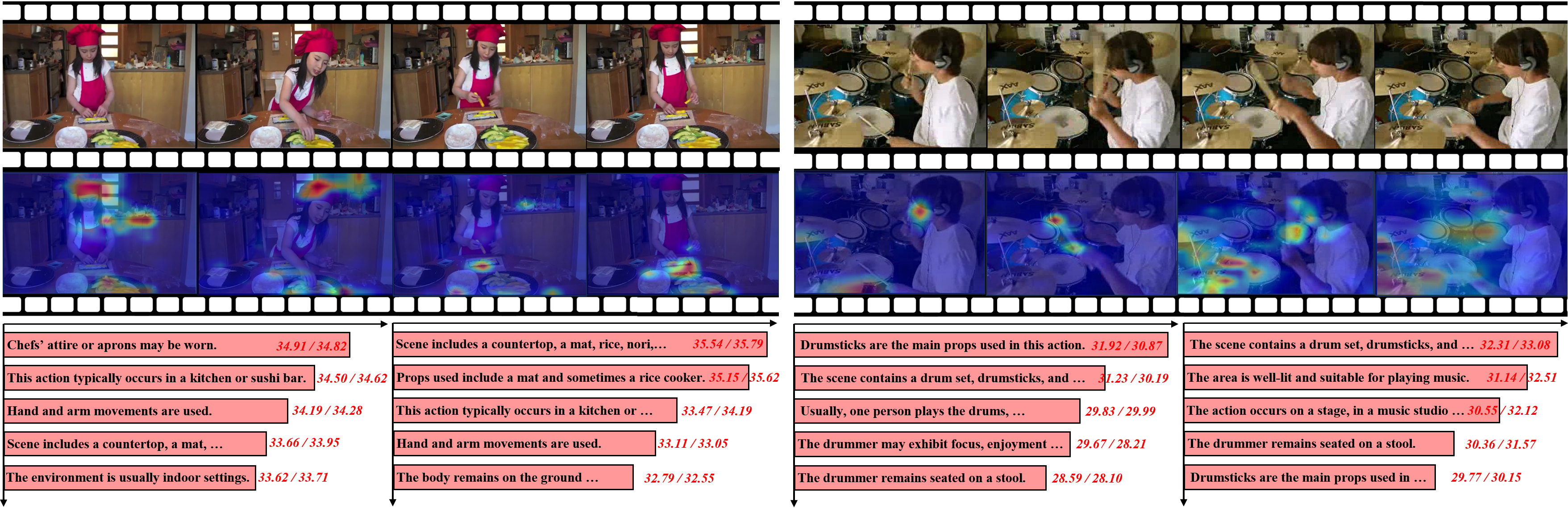}
  \vspace{-0.1cm}
  \caption{The second row illustrates the visualization of visual attention maps from corresponding video frames. The third row displays the top-5 prompts for each video frame. For visual simplicity, two video frames are grouped together as they share the same top 5 prompts. The corresponding CLIP match scores are shown to the right of the bar graphs.}
  \label{fig:case}
\end{figure*}
\setlength{\belowcaptionskip}{-0.0cm}

\subsection{Ablation Experiments}
In this section, extensive ablation experiments are conducted to demonstrate the efficacy of our AP-CLIP's components. We start from ViFi-CLIP \cite{rasheed2023fine} as the baseline, employing ViT-B/16 as its backbone. The model is pre-trained on the Kinetics-400 dataset. Evaluation is carried out across various datasets, including HMDB-51, UCF-101, and Kinetics-600, to assess the impact of \textit{action-conditioned prompts} and the \textit{multi-modal action knowledge alignment} within our framework.

\vspace{0.05cm}
\noindent\textbf{{(\romannumeral 1) Is the ``\textit{action-conditioned prompts}" important? }}
\vspace{0.05cm}

Table \ref{tab:ablation_prompt} assesses the effectiveness of different prompting strategies on model performance. These include the single prompt from ViFi-CLIP~\cite{rasheed2023fine}, a collection of manually crafted prompts in Open-VCLIP~\cite{weng2023transforming}, and several approaches involving prompts generated by GPT. The generative strategies consist of Customized prompts in \cite{pratt2023does}, action prompts directly generated by LLMs, and our specifically designed prompts. The findings suggest that LLM-generated prompts, with their rich knowledge base, not only reduce manual labor but also substantially bolster performance. 
Moreover, while directly utilizing the extensive knowledge from LLMs is a trivial solution, the hierarchical generation strategy inherent in our action-conditioned prompt generation can better manage knowledge from various perspectives, resulting in better generalizability. This approach better mirrors human cognitive processes, crafting prompts that more effectively aid in action recognition. We present additional ablation studies in the appendix, including the impact of different LLM-prompts and attributes.

\vspace{0.05cm}
\noindent\textbf{{(\romannumeral 2) Is the ``\textit{multi-modal action knowledge alignment}" important? }} 
\vspace{0.05cm}

The multi-modal action knowledge alignment mechanism (MAKA) aims to align multi-attribute prompts with videos and is thus applicable exclusively to methods that utilize multiple prompts. Table \ref{tab:ablation_maka} showcases that the incorporation of MAKA with both Customized prompts and our Action-conditioned Prompts results in uniform performance gains. This suggests that the alignment of multi-modal knowledge facilitates the model's development of a more comprehensive and nuanced comprehension of actions, thereby enhancing recognition performance.

\subsection{Interpretability}

Owing to the multi-attribute action-conditioned prompts and the multi-modal action knowledge alignment, our approach exhibits notable interpretability, which aids in elucidating the rationale behind the model's judgments. In Figure \ref{fig:case}, we present the attention map visualizations of video frames alongside the score distributions for various attribute prompts. These visualizations reveal that different frames within a video garner varying focal points, which correspond to distinct prompts. For instance, as depicted in Figure \ref{fig:case} (left), initial frames focus primarily on the actor's clothing and the surrounding environment, where actor-related and scene-related prompts provide more clues for judgment. Conversely, subsequent frames shift attention to scene elements and props. Similarly, the example on the right also demonstrates a shift in focus, from props to the environment, and illustrates how well these aspects match with the corresponding prompts.
These observations affirm that a model's access to diverse knowledge enhances its action recognition abilities. Crucially, they also highlight the importance of concept alignment within the video to corresponding prompts, a vital factor for interpretability. In the appendix, we provide a more detailed visualization of the frame-to-prompt correspondence.


\setlength{\belowcaptionskip}{-0.1cm}
\begin{table}[!ht]
\centering
\setlength{\tabcolsep}{1mm}{
\caption{Analysis on different prompting strategies.}
\vspace{-0.15cm}
\label{tab:ablation_prompt}
\resizebox{0.90\columnwidth}{!}{
\begin{tabular}{lllll}
\toprule
\rowcolor{tabhighlight} Method   & HMDB-51 & UCF-101 & K-600\\
\midrule
Single~\cite{rasheed2023fine}     	& 51.3 &	76.8 &	71.2	\\
Set~\cite{weng2023transforming}  &	52.9 &	79.1 &	71.4 \\
Customized~\cite{pratt2023does} &	{53.1} &	 {78.8} &	 {71.8} \\
\midrule

LLMs-Direct &	{52.7} &	 {77.9} &	 {71.1} \\
AP(ours)  &	\textbf{54.3} \textcolor{MidnightBlue}{{(+1.6)}} &	\textbf{81.5} \textcolor{MidnightBlue}{{(+3.6)}} &	\textbf{72.9} \textcolor{MidnightBlue}{{(+1.8)}}\\
\bottomrule
\end{tabular}
}}
\vspace{-0.2in}
\end{table}

\begin{table}[!ht]
\centering
\setlength{\tabcolsep}{1mm}{
\caption{Ablation of ``\textit{multi-modal action knowledge alignment}".}
\vspace{-0.15cm}
\label{tab:ablation_maka}
\resizebox{0.90\columnwidth}{!}{
\begin{tabular}{llll}
\toprule
\rowcolor{tabhighlight} Method   & HMDB-51 & UCF-101 & K-600\\
\midrule
Customized~\cite{pratt2023does} &	{53.1} &	 {78.8} &	 {71.8} \\
+ MAKA &	53.9 \textcolor{MidnightBlue}{{(+0.8)}}  &	 79.9 \textcolor{MidnightBlue}{{(+1.1)}}  &	 72.2 \textcolor{MidnightBlue}{{(+0.4)}}  \\
\midrule
AP(ours)  &	54.3 &	81.5 &	72.9\\
+ MAKA  &	\textbf{55.4} \textcolor{MidnightBlue}{{(+1.1)}} &	\textbf{82.4} \textcolor{MidnightBlue}{{(+0.9)}} &	\textbf{73.4} \textcolor{MidnightBlue}{{(+0.5)}}\\
\bottomrule
\end{tabular}
}}
\vspace{-0.2in}
\end{table}

\section{Conclusion}
\label{sec:con}


In this work, we blend video models with Large Language Models (LLMs) to enhance open-vocabulary action recognition. Our strategy centers on generating Action-conditioned Prompts that enrich the textual embeddings in the CLIP model with human prior knowledge. Building on these knowledge-based prompts, we introduce a multi-modal action knowledge alignment mechanism to align concepts in video and knowledge encapsulated within the prompts. Extensive experiments not only demonstrate the effectiveness of our approach but also highlight its superior interpretability. By highlighting the significance of knowledge-based prompting, we anticipate our research will inspire further exploration and innovation in this field.

{
    \small
    \bibliographystyle{ieeenat_fullname}
    \bibliography{main}
}
\appendix

\setcounter{section}{0}
\clearpage
\setcounter{page}{1}
\setcounter{table}{0}
\setcounter{figure}{0}
\maketitlesupplementary


This supplementary material offers extensive additional details and more qualitative and quantitative analysis complementing the main paper. The content is organized as follows:

\begin{itemize}
    \item Prompts for LLMs and Responses
    (Appendix~\ref{appendix:prompts})
    \item Ablation on LLM-prompts and attributes (Appendix~\ref{appendix:llm_prompts})
    \item Additional qualitative visualization (Appendix~\ref{appendix:vis})
    \item More details of the training objectives and procedures (Appendix~\ref{appendix:objectives})
    \item More implementation details (Appendix~\ref{appendix:implementation_details})
    \item More details of experimental settings (Appendix~\ref{appendix:experiment_settings}).
    \item More details of datasets (Appendix~\ref{appendix:datasets}).
    
\end{itemize}

\section{Prompts for LLMs and Responses}
\label{appendix:prompts}

Drawing inspiration from LLMs’ chain-of-thought \cite{maniparambil2023enhancing,wei2022chain}, we
adopt a hierarchical generation strategy to generate multi-
attribute Action-conditioned Prompts. Initially, we prompt the LLM to enumerate a list of attributes crucial for distinguishing actions. Subsequently, the second prompt directs the LLM to formulate descriptions utilizing the attributes identified in the first step, while specifying the desired output format. We prefer a JSON dictionary format for the output, comprising a list of sentences, each linked to a specific attribute. This format is not only straightforward to utilize but also maintains the granularity of attribute-level information, thereby promoting a richness in attributes.

\subsection{Which attributes are critical for action recognition?}

Example of the first prompt for determining attributes:

\begin{quote}\textit{ \textbf{Q:} ~
I am currently utilizing CLIP for video action classification tasks. I want to create action class attributes for a zero-shot video action recognition algorithm to classify different actions. List some attributes that take into account the scene of the action's occurrence, information about the actor performing the action, and details of the actor's bodily movements, which can contribute to a class description that aids in distinguishing actions.}
\end{quote}

GPT-4 response:

\begin{quote}
    \textit{ \textbf{A:}~
 For creating effective action class attributes for zero-shot video action recognition using a tool like CLIP, it's crucial to integrate aspects that cover the scene, the actor, and the nature of the movements. Here are some attributes that can be considered for each category: ...
    }
\end{quote}

In Table \ref{table:prompts}, we display the list of attributes generated by GPT. After removing duplicates and unimportant content, we finalized a set of 12 core attributes. These are distributed across the three components and are illustrated in Figure \ref{fig:overview1} in the main paper.

\begin{table*}[t]
\resizebox{\textwidth}{!}{%
\begin{NiceTabular}{l|l|l}
\toprule
\textbf{Type} & \textbf{Attributes}       & \textbf{Explanation}                                                    \\
\midrule
\rule{0pt}{2ex}
Scene-Related    &    Scene Type * & Soccer field, Kitchen, Street, Gymnasium, etc.   \\

    &    Scene Elements * & Soccer ball, Goalpost, Stove, Traffic sign, etc.   \\  
    &    Scene Conditions * & Sunny, Rainy, Indoor, Outdoor, etc.   \\
    &    Prop Usage * & Soccer ball, Knife, Cookware, etc.   \\
    &    Weather Conditions & Sunny, cloudy, rainy, snowy, foggy, etc. \\
    & Human Crowds & Busy streets, empty spaces, group gatherings, etc. \\
    & Specific Locations & Parks, offices, classrooms, industrial areas, etc. \\
    & Terrain Type & Flat ground, hilly area, uneven surfaces, water bodies, etc.\\
    & Cultural Context & Specific to a region or community.\\
    &Color and Texture& Bright, dark, colorful, monochrome environments, etc.\\

Actor-Related    &    Number of Actors * & Single, Double, Multiple.  \\
&    Clothing * & Sportswear, Chef's uniform, Police uniform, etc.  \\
&    Actor Identity * & Athlete, Chef, Policeman, etc.  \\
&    Facial Expression * & Happy, Sad, Angry, Surprised, etc.  \\
&Age Group  & Children, teenagers, adults, elderly. \\
& Clothing Style & Formal, casual, athletic, traditional. \\
& Emotional State & Stressed, calm, excited, bored.\\
&Hairstyle and Accessories& Short, long hair, hats, glasses.\\
&Visible Health Conditions& Signs of fatigue, injury, robust health.\\
&Ethnicity or Cultural Background& Diverse cultural representations.\\

Body-Related & Body Move Speed * & Fast, Medium, Slow, etc.\\
&Body Part Movement *  & Hand, Leg, Head, etc.\\
& Body Posture *  & Standing, Sitting, Lying, Bending, etc.\\
&Body Position *  &  In contact with ground, Off the ground, etc.\\
&Purpose of Body Movement & Functional, expressive, recreational, competitive, etc.\\
&Changes in Posture& Standing, sitting, lying, bending, etc.\\

&Movement Complexity& Simple, complex, repetitive, unique, etc.\\
&Body Coordination Level& Coordinated, uncoordinated, synchronized, etc.\\
&Body Movement Style & Graceful, abrupt, fluid, stiff, etc.\\  
&Body Rhythm and Timing& Regular, irregular, rhythmic, sporadic, etc.\\

\end{NiceTabular}
}%
\caption{List of attributes and their corresponding explanations as provided by GPT responses. Attributes marked with an asterisk (*) are those that were ultimately selected.}
\label{table:prompts}
\end{table*}

\subsection{Describe the action about critical attributes}
\label{sec:prompts}
We then construct a set of LLM-prompts. Their purpose is to inquire specifically about certain actions in relation to identified attributes, aiming to generate standardized, knowledge-enriched descriptive sentences. Examples of LLM-prompts are illustrated as follows:

\begin{quote}\textit{ \textbf{LLM-prompt1:}
Describe the following actions by adding one sentence about its \{attribute\} for \{action\}.
}
\end{quote}
\begin{quote}\textit{ \textbf{LLM-prompt2:}
Visually describe \{action\} about its \{attribute\} in one sentence.
}
\end{quote}
\begin{quote}\textit{ \textbf{LLM-prompt3:}
What are the primary characteristics of \{action\} interm of its \{attribute\}, explain in one sentence.
}
\end{quote}
\begin{quote}\textit{ \textbf{LLM-prompt4:}
Explain how \{action\} typically manifests in terms of its \{attribute\}, providing a descriptive sentence.
}
\end{quote}
\begin{quote}\textit{ \textbf{LLM-prompt5:}
Analyze \{action\} in the context of its \{attribute\}, and describe it in one sentence.}
\end{quote}

Then, for the selected LLM-prompt, we generate a suite of 12 distinct action-conditioned prompts, ensuring that every action is matched with tailored descriptive phrases, as depicted in Figure \ref{fig:overview1} in the main paper. Additionally, we instructed GPT to output the results in a JSON format, where each key is an attribute and the corresponding value is the descriptive sentence for that attribute. 

Using \textit{Making Sushi, Drumming} and \textbf{LLM-prompt1} as an example, GPT's output response is as follows:

\begin{quote}\textit{ ``Making Sushi": \{
    ``Scene Type": ``This action typically occurs in a kitchen or sushi bar.",
    ``Scene Elements": ``The scene includes a countertop, a bamboo sushi mat, rice, nori (seaweed sheets), and various fillings like fish, vegetables, and condiments.",
    ``Scene Conditions": ``The environment is usually controlled, indoor settings.",
    ``The usage of props in the action": ``Props used include a bamboo sushi mat, a sharp knife, and  a rice cooker.",
    ``Number of Actors": ``This action can be performed by a single individual or with others, especially in a professional setting.",
    ``Clothing": ``Chefs’ attire or aprons may be worn.",
    ``Actor Identity": ``Anyone who enjoys cooking or is employed as a sushi chef can perform this action.",
    ``Facial Expression": ``Expressions can range from focused to satisfied, especially when creating or enjoying a beautiful sushi roll.",
    ``Body move Speed": ``The action is performed at a moderate to slow pace, allowing for precision in rolling the sushi.",
    ``Body Part Movement": ``Hand and arm movements are used to spread the rice, place the fillings, and roll the sushi using the bamboo mat.",
    ``Body Posture": ``Individuals are typically standing while performing this action.",
    ``Body Position": ``The body remains on the ground during this action." 
    \}, \\
    ``Drumming": \{
    ``Scene Type": ``The action typically occurs on a stage, in a music studio, or in a practice room.",
    ``Scene Elements": ``The scene contains a drum set, drumsticks, and possibly other musical instruments and musicians.",
    ``Scene Conditions": ``The area is well-lit and acoustically suitable for playing music.",
    ``The usage of props in the action": ``Drumsticks and a drum set are the main props used in this action.",
    ``Number of Actors": ``Usually, one person plays the drums, although other musicians may be present.",
    ``Clothing": ``The drummer wears casual or performance attire, depending on the setting.",
    ``Actor Identity": ``The actor is a drummer, possibly part of a band or ensemble.",    
    ``Facial Expression": ``The drummer may exhibit focus, enjoyment, and rhythm as they play.",
    ``Body move Speed": ``The action varies in speed, with fast, rhythmic drumming or slower, deliberate strikes.",
    ``Body Part Movement": ``The drummer's arms move rapidly to strike the drums, while the feet operate the bass drum and hi-hat pedals.",
    ``Body Posture": ``The body posture is seated with a straight back, and arms and legs in motion.",
    ``Body Position": ``The drummer remains seated on a stool during the action." 
    \}
}
\end{quote}

\textbf{The complete prompts will be made publicly available promptly after the paper is accepted.}

\section{Ablation on LLM-prompts and attributes.}
\label{appendix:llm_prompts}
We present additional ablation studies including the impact of different LLM-prompts and attributes.

\setlength{\belowcaptionskip}{-0.1cm}
\begin{table}[t]
\centering
\setlength{\tabcolsep}{1mm}{
\caption{The impact of different LLM-prompts.}
\vspace{-0.15cm}
\label{tab:ablation_llm_prompt}
\resizebox{0.95\columnwidth}{!}{
\begin{tabular}{cccc}
\toprule
\rowcolor{tabhighlight} Method   & HMDB-51 & UCF-101 & K-600\\
\midrule
LLM-prompt1     	& 54.7 &	\textbf{81.1} &	\textbf{72.4}	\\
LLM-prompt2  & 54.3 &	80.8 &	72.2	\\
LLM-prompt3  & 54.5 &	81.0 &	72.1	\\
LLM-prompt4  & \textbf{54.8} &	80.5 &	72.3	\\
LLM-prompt5  & 54.1 &	80.7 &	71.8	\\
\midrule
LLM-prompt num 2 &	55.1 &	81.9 &	73.3 \\
\rowcolor{tabhighlight}LLM-prompt num 3 &	\textbf{55.4} &	82.4 &	\textbf{73.4} \\
LLM-prompt num 4 &	55.3 &	\textbf{82.7} &	\textbf{73.4} \\
LLM-prompt num 5 &	55.2 &	82.6 &	73.7 \\

\bottomrule
\end{tabular}
}}
\end{table}

\begin{table}[t]
\centering
\setlength{\tabcolsep}{1mm}{
\caption{The impact of different attributes.}
\label{tab:ablation_attributes}
\resizebox{0.95\columnwidth}{!}{
\begin{tabular}{cccc}
\toprule
\rowcolor{tabhighlight} Method   & HMDB-51 & UCF-101 & K-600\\
\midrule
Scene     	& 52.5 &	82.0 &	72.5	\\
Actor  &	54.1 &	81.4 &	72.9 \\
Body  &	54.8 &	81.3 &	72.5 \\
Scene+Actor  &	54.2 &	82.3 &	73.1 \\
Scene+Body  &	54.5 &	82.1 &	73.1 \\
Actor+Body &  55.1	 &	82.1 &	73.3 \\
\rowcolor{tabhighlight} Scene+Actor+Body &	\textbf{55.4} &	\textbf{82.4} &	\textbf{73.4} \\
\bottomrule
\end{tabular}
}}
\vspace{-0.2in}
\end{table}

\subsection{The impact of different LLM-prompts}
Table \ref{tab:ablation_llm_prompt} showcases the impact of different LLM-prompts on performance. The upper section presents results using individual LLM-prompt, while the lower section shows the collective performance when selecting the best-performing set of \{num\} prompts. It is observable that variations in LLM-prompts have a minimal effect on performance (less than 0.6\%), underscoring their robustness compared to manually designed prompts fed into CLIP. Furthermore, while integrating multiple LLM-prompts can enhance performance, the gains become marginal as the number increases. Considering the trade-off between efficiency and performance, we default to employing a set of three LLM-prompts, including \textit{LLM-prompt1}, \textit{LLM-prompt3}, and \textit{LLM-prompt4}.

\subsection{The impact of different attributes}
Table \ref{tab:ablation_attributes} illustrates the performance impact of prompts associated with different attributes across various datasets. It is evident that prompts tied to specific attributes can enhance model performance. However, the degree of improvement attributed to each category varies depending on the dataset. For instance, prompts related to ``Body'' show a greater benefit for HMDB-51, while 'Actor' prompts yield more substantial gains for UCF-101 and Kinetics-600, indicating a divergence in focal points among datasets. Overall, prompts that amalgamate all three aspects, \ie, Scene, Actor, and Body, achieve a comprehensive performance boost across datasets, which mitigates dataset biases. This underlines the necessity of our hierarchical generation approach in producing multi-attribute prompts and its effectiveness in addressing the varying scenarios of different datasets.

\section{Additional qualitative visualization}
\label{appendix:vis}

\begin{figure*}[t]
  \centering
  \includegraphics[width=0.90\textwidth]{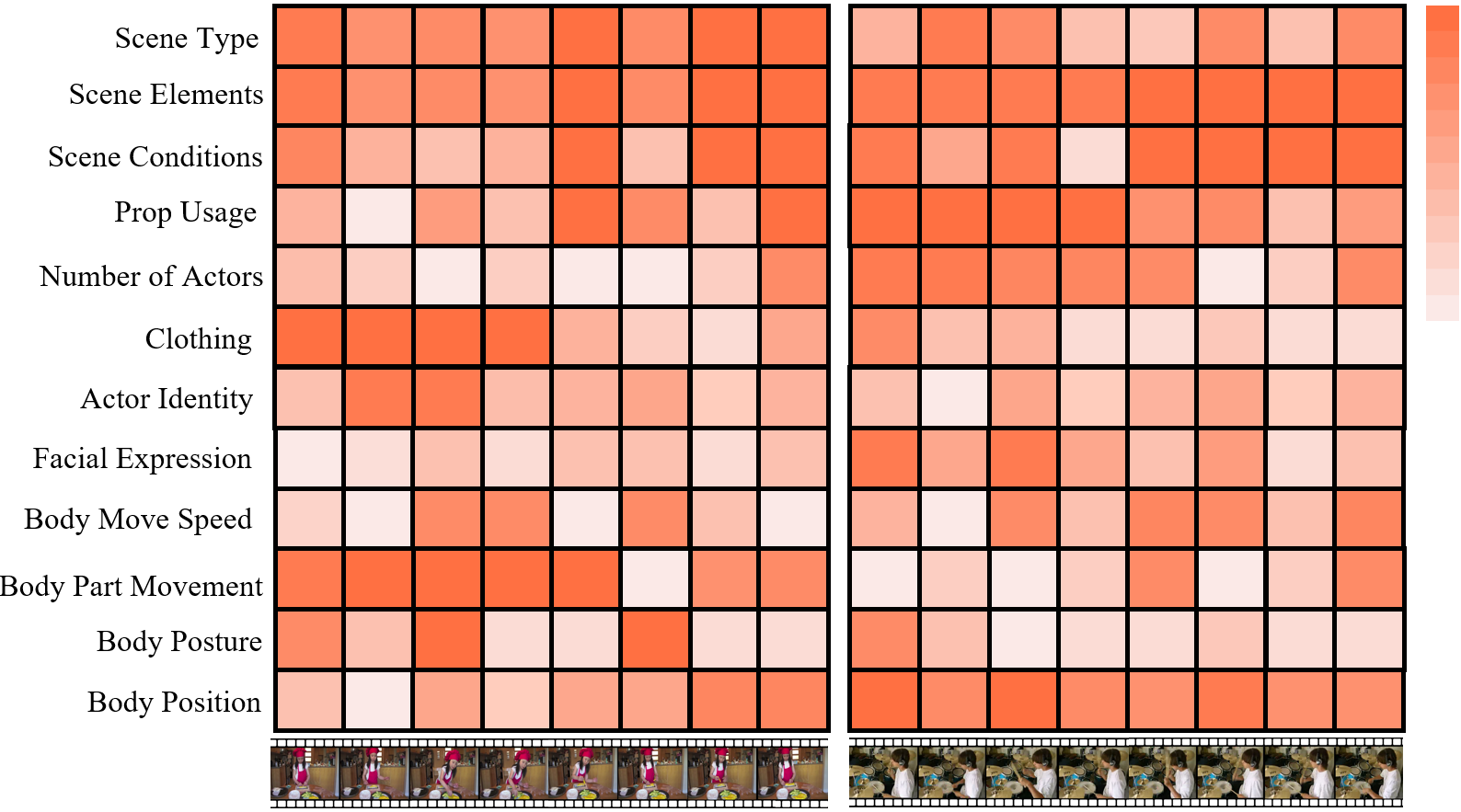}
  \vspace{-0.1cm}
  \caption{Illustrative heatmap of the association strengths between 8 video frames and a set of 12 prompts exemplified for the actions ``\textit{Making Sushi}" and ``\textit{Drumming}", as detailed in Appendix \ref{sec:prompts}. The intensity of the color corresponds to the association strength, calculated using the CLIP match score. The heatmap reveals that different frames are associated with prompts from varying attributes, providing a clear pathway to understanding how the model discerns actions through both visual and textual cues.}
  \label{fig:app}
\end{figure*}

We provide a more detailed visualization of the frame-to-prompt correspondence in Figure \ref{fig:app}. 

\setlength{\belowcaptionskip}{-0.0cm}

\section{More details of the training objectives and procedures}
\label{appendix:objectives}
Given a video $V \in \mathbb{R}^{T \times H \times W \times 3} $ with $T$ frames and a set of prompts  $C$, where $V$ and $C$ are sampled from a set of videos $\mathcal{V}$ and a collection of action category $\mathcal{C}$ respectively, we feed the $T$ frames into the video encoder $f_{\theta_v}$ and the text $C$ into the text encoder $f_{\theta_t}$ to obtain a video representation $v \in \mathbb{R}^{n_v \times d}$ and text embeddings $c \in \mathbb{R}^{n_t \times d}$ correspondingly.

For a batch of videos, the similarity $sim(.)$, between all the video representation $v$ and the corresponding text embeddings $c^{\prime}$ is maximized to fine-tune the CLIP model via cross-entropy (CE) objective with a temperature parameter $\tau$,
\begin{align*}
\Lagr = -\sum_{v \sim \mathcal{V}} \log\frac{\text{exp}(sim(v,c^{\prime})/\tau)}{\sum_{c \sim \mathcal{C}}\text{exp}(sim(v, c)/\tau)}, 
\end{align*}
where $c^{\prime}$ represents the ground truth action-conditioned prompts corresponding to video $v$. We employ Equation \eqref{equ:sim_f} in the main paper to compute the fine-grained similarity between prompts and each video.

We also provide a PyTorch-style pseudocode for the \textit{multi-modal action knowledge alignment} mechanism in Algorithm \ref{alg:pytorchmask} to aid in understanding the entire alignment procedure.

\begin{algorithm}[tb]
\caption{PyTorch-style pseudocode for multi-modal action knowledge alignment mechanism.}
\label{alg:pytorchmask}
\definecolor{codeblue}{rgb}{0.25,0.5,0.5}
\definecolor{deepred}{rgb}{0.95,0,0}
\lstset{
      commentstyle=\fontsize{7.2pt}{7.2pt}\color{codeblue},
      keywordstyle=\fontsize{7.2pt}{7.2pt},
      emphstyle=\color{red},    %
    }
\newcommand*{\ttfamilywithbold}{\fontsize{7.2pt}{7.2pt}\fontfamily{lmtt}\selectfont}
\begin{lstlisting}[basicstyle=\ttfamilywithbold, language=python]
# e_c: Action prompts embeddings              
# e_v: videos embeddings
# f_t: text encoder network     
# f_v: video encoder network
# B: batch size                 
# D: dimensionality of the embeddings
# K:  number of categories  
# N_v: number of frames        
# N_t: number of prompts for each action

def action_knowledge_alignment(C, V):    
    # compute embeddings
    e_c = f_t(C) # KxN_txD
    e_v = f_v(V) # BxN_vxD
    
    # normalize representation
    e_c = e_c / e_c.norm(dim=-1, keepdim=True)  
    # KxN_txD
    e_v = e_v / e_v.norm(dim=-1, keepdim=True)  
    # BxN_vxD
    
    # fine-grained relevancy between prompts and frames 
    logits =  torch.einsum('bvd,ktd->bktv', [e_v, e_c]) 
    # BxKxN_txN_v
    
    
    t2v_logits, t2v_max_idx = logits.max(dim=-1)  
    # BxKxN_txN_v -> BxKxN_t
    t2v_logits = t2v_logits.mean(dim=-1) 
    # BxKxN_t -> BxK
    v2t_logits, v2t_max_idx = logits.max(dim=-2)  
    # BxKxN_txN_v -> BxKxN_v
    v2t_logits = v2t_logits.mean(dim=-1)  
    # BxKxN_v-> BxK
    
    alignment_logits = (t2v_logits + v2t_logits) / 2.0  
    # BxK

    return alignment_logits
\end{lstlisting}
\end{algorithm}

\section{More implementation details}
\label{appendix:implementation_details}
Our adaptation of the CLIP model follows \cite{rasheed2023fine}, with tailored modifications to the prompts and fine-grained similarity function used. We preprocess all frames to a uniform spatial dimension of 224×224 pixels. Optimization is carried out using an AdamW optimizer with a weight decay set at 0.001. Adaptations in epochs, batch size, and learning rate are made to suit varying experimental conditions, as outlined subsequently. For the \textit{zero-shot setting}, CLIP is trained on the Kinetics-400 dataset for 10 epochs, utilizing a batch size of 256 and a learning rate of 8e-6. In both the \textit{base-to-novel generalization} and \textit{few-shot settings}, training proceeds in a few-shot manner with a batch size of 64 and a learning rate of 2e-6. Under the \textit{fully-supervised setting}, we extend CLIP's training on Kinetics-400 to 30 epochs, with an increased batch size of 256 and a learning rate of 22e-6. These experiments were performed on a computing cluster equipped with 8 A100 GPUs.

As our method adopts action-conditioned prompts, it diverges from the ViFI approach which utilizes learnable prompting methods. For experiments transitioning from utilizing parameters pre-trained on Kinetics-400 to the base-to-novel and few-shot scenarios, we fine-tune the pre-trained CLIP model directly, with a batch size of 64 and a learning rate of 2e-6. Empirical evidence from our experiments corroborates the effectiveness of this approach, as shown in Table \ref{base_novel} and \ref{p1_few_shot}  in the main paper.

\section{More details of experimental settings}
\label{appendix:experiment_settings}
We align with previous methods \cite{wang2021actionclip,ni2022expanding,ju2022prompting,rasheed2023fine} for various settings including zero-shot, base-to-novel, few-shot, and fully-supervised. Specifically, we utilize 8 frames and employ multi-view inference incorporating 2 spatial crops and 2 temporal views. In the fully supervised setting, our approach extends to using 16 frames, combined with multi-view inference featuring 4 spatial crops and 3 temporal views, consistent with compared methods. Each sampled frame is spatially scaled on the shorter side to 256, with a center crop of 224.

\vspace{0.1in}
\noindent \textbf{Zero-shot setting}: In the zero-shot setting, models trained on the Kinetics-400 dataset undergo testing on three distinct datasets: HMDB-51, UCF-101, and Kinetics-600. For HMDB-51 and UCF-101, performance is assessed across the three standard validation splits, with the top-1 average accuracy being reported. Regarding Kinetics-600, following the methodology of ~\cite{chen2021elaborative}, the evaluation focuses on the 220 categories that do not overlap with those in Kinetics-400. Here, we also document top-1 average accuracies derived from three randomly generated splits, each inclusive of 160 categories. This assessment utilizes a multi-view strategy, encompassing 2 different spatial crops and 2 temporal clips, amounting to a total of 32 frames.

\vspace{0.1in}
\noindent \textbf{Base-to-novel setting}: Following \cite{rasheed2023fine}, we employ a \emph{base-to-novel generalization} setting for extensive analysis on the generalization ability of various approaches. In this setting, models undergo initial training on a set of 'base' (seen) classes using a few-shot approach and are then evaluated on a set of 'novel' (unseen) classes. Our analysis spans four datasets: Kinetics-400, HMDB-51, UCF-101, and SSv2 as \cite{rasheed2023fine}. For each, we employ three training splits, with 16 shots per action category, selected at random. Categories are divided into two equal groups: the more frequently occurring actions serve as 'base' classes, while the less common ones are designated as 'novel' classes. Evaluations are performed on the respective validation splits, with HMDB-51 and UCF-101 limited to their first split, and full validation splits used for Kinetics and SSv2. The setting also follows a multi-view strategy, integrating two spatial crops and two temporal clips. 

\vspace{0.05in}
\noindent \textbf{Few-shot setting}: In the few-shot scenario, we establish a general K-shot configuration by randomly selecting K examples from each category for training purposes. Concretely, for the datasets HMDB-51, UCF-101, and SSv2, we utilize 2, 4, 8, and 16 shots. Performance evaluations for HMDB-51 and UCF-101 are conducted using their first validation split, while for SSv2, the entire validation split is used. This setting also employs a multi-view strategy, integrating two spatial crops and two temporal clips.

\vspace{0.1in}
\noindent \textbf{Fully-supervised setting}: In the fully-supervised setting, models trained on the Kinetics-400 dataset are assessed against its entire validation set. We conduct evaluations using 16 frames and employ a multi-view inference approach, which includes three distinct spatial crops and four temporal segments.

\section{More details of dataset}
\label{appendix:datasets}

\noindent  We conduct our analysis on five established action recognition benchmarks: Kinetics-400~\cite{k400} and Kinetics-600~\cite{k600}, HMDB-51~\cite{kuehne2011hmdb}, UCF-101~\cite{soomro2012ucf101} and Something-Something v2 (SSv2)~\cite{goyal2017ssv2}. 

\vspace{0.1in}
\noindent \textbf{Kinetics-400 and Kinetics-600}: The Kinetics-400 and Kinetics-600 datasets are comprehensive collections designed for human action recognition, containing approximately 240k training and 20k validation videos across 400 action classes, and around 410k training and 29k validation videos covering 600 action classes, respectively. Originating from diverse YouTube videos, each clip is roughly 10 seconds in length, capturing a concise action moment. Kinetics-600 builds upon the foundation set by Kinetics-400, introducing an additional 220 action categories that enrich the dataset, particularly for evaluating zero-shot learning capabilities. While these datasets offer a wide variety in content, it's noteworthy that there is a tendency towards spatial appearance biases. These extensive collections present an opportunity for models to demonstrate their proficiency in recognizing a broad spectrum of human activities.

\vspace{0.1in}
\noindent \textbf{HMDB-51}: The HMDB-51 dataset comprises 6,849 video clips distributed across 51 distinct action categories, ensuring a minimum of 101 clips per category. This dataset has been amassed from various realistic sources and is designed for a balanced evaluation. Officially, it offers three different training/testing splits. To maintain uniformity across categories, each split is configured to include 70 training and 30 test samples per category, while leaving 1,746 videos as 'unused' to preserve sample balance. This structure of training and testing allows for a consistent and fair assessment of the model's performance across the full spectrum of actions.

\vspace{0.1in}
\noindent \textbf{UCF-101}: 
The UCF-101 dataset is a benchmark for human action recognition featuring 13,320 video clips sourced from YouTube, spanning 101 action categories. These categories encompass a broad range of actions including human-object interaction, body motion, human-human interaction, playing musical instruments, and various sports. Each video is a succinct representation of an action, averaging 7.21 seconds in length, derived from realistic scenarios. For evaluation consistency, the dataset is divided into three standard splits, with the official split allocating 9,537 videos for training and 3,783 for testing.

\vspace{0.1in}
\noindent \textbf{Something-Something v2 (SSv2)}: The SSv2 dataset is a comprehensive video action recognition benchmark that specifically emphasizes temporal modeling. It features 220,487 videos across 174 action categories, capturing humans interacting with everyday objects. The actions depicted in SSv2 are finely detailed, focusing on nuanced activities such as covering or uncovering objects, thereby showcasing a dataset with a strong temporal bias distinct from other datasets like K400. The videos range from 2 to 6 seconds in length, highlighting the rich temporal details over static scenes. The standard dataset split includes 168,913 training videos and 24,777 validation videos. We evaluate and report the top-1 accuracy using the validation split. SSv2 uniquely prioritizes dynamic information in videos over static scene contexts, presenting a challenging environment for models to accurately capture and interpret temporal action dynamics.

\end{document}